\documentclass[letterpaper, 10 pt, conference]{ieeeconf}  

\pdfoutput=1

\IEEEoverridecommandlockouts                              

\usepackage{hyperref}       
\usepackage{url}            
\usepackage{booktabs}       
\usepackage{amsfonts}       
\usepackage{nicefrac}       
\usepackage{microtype}      
\usepackage[dvipsnames]{xcolor}
\usepackage{graphicx}
\usepackage[ruled,vlined]{algorithm2e}
\usepackage{bm}
\usepackage{amsmath}
\usepackage{multirow}
\usepackage{threeparttable}
\usepackage{adjustbox}
\usepackage{eqparbox}
\usepackage{arydshln}
\usepackage{wrapfig}

\usepackage{graphicx}
\usepackage{amsmath,amssymb} 
\usepackage{tabularx}
\usepackage{multirow}
\usepackage{subfig}

\DeclareMathOperator*{\argmin}{arg\,min}

\title{\LARGE \bf
Unsupervised Domain Adaptation in LiDAR Semantic Segmentation with Self-Supervision and Gated Adapters
}

\author{Mrigank Rochan$^{*}$, Shubhra Aich$^{*}$, Eduardo R. Corral-Soto, Amir Nabatchian, and Bingbing Liu
\thanks{*Equal contribution.}
\thanks{Work done while all the authors were with Huawei Noah's Ark Lab. Corresponding author is Bingbing Liu ({\tt\small liu.bingbing@huawei.com}).}%
}

\begin{document}

\maketitle
\thispagestyle{empty}
\pagestyle{empty}

\begin{abstract}
In this paper, we focus on a less explored, but more realistic and complex problem of domain adaptation in LiDAR semantic segmentation. There is a significant drop in performance of an existing segmentation model when training (source domain) and testing (target domain) data originate from different LiDAR sensors. To overcome this shortcoming, we propose an unsupervised domain adaptation framework that leverages unlabeled target domain data for self-supervision, coupled with an unpaired mask transfer strategy to mitigate the impact of domain shifts. Furthermore, we introduce the gated adapter module with a small number of parameters into the network to account for target domain-specific information. Experiments adapting from both real-to-real and synthetic-to-real LiDAR semantic segmentation benchmarks demonstrate the significant improvement over prior arts.

\end{abstract}

\section{Introduction}

In autonomous driving systems, 3D semantic segmentation plays an indispensable role since it provides precise and robust perception of the surrounding environment. For 3D perception, LiDAR (Light Detection and Ranging) is a commonly used sensor that delivers accurate distance measurements of the encompassing 3D world. As a consequence, LiDAR-based perception has been receiving a lot of scientific interest. 

\begin{figure} 
	\centering
	\begin{tabular}{cc}
		\subfloat[]{
			\includegraphics[width=0.45\columnwidth, clip]{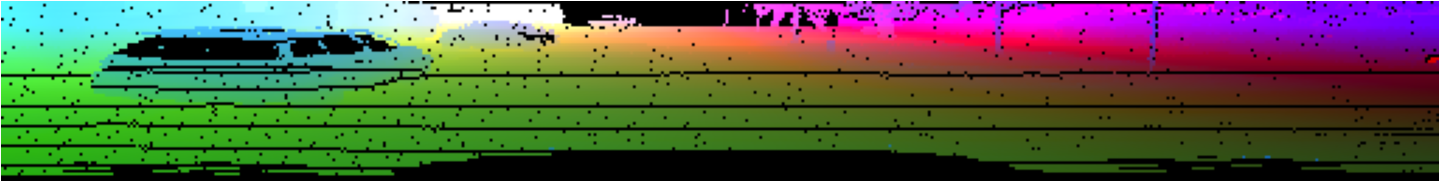} 
		}	
		\subfloat[]{
			\includegraphics[width=0.45\columnwidth, clip]{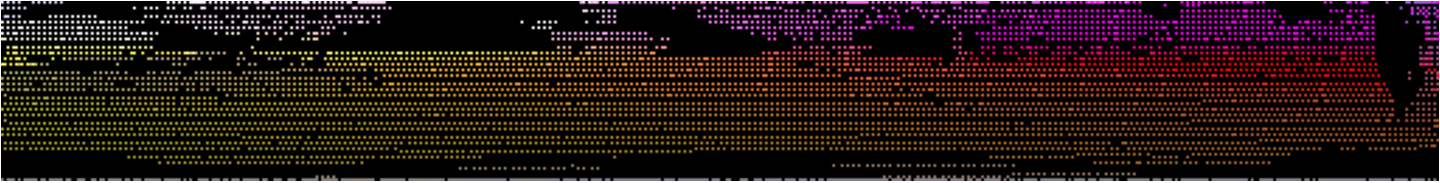}  
		}\\	
		\subfloat[]{
			\includegraphics[width=0.45\columnwidth, clip]{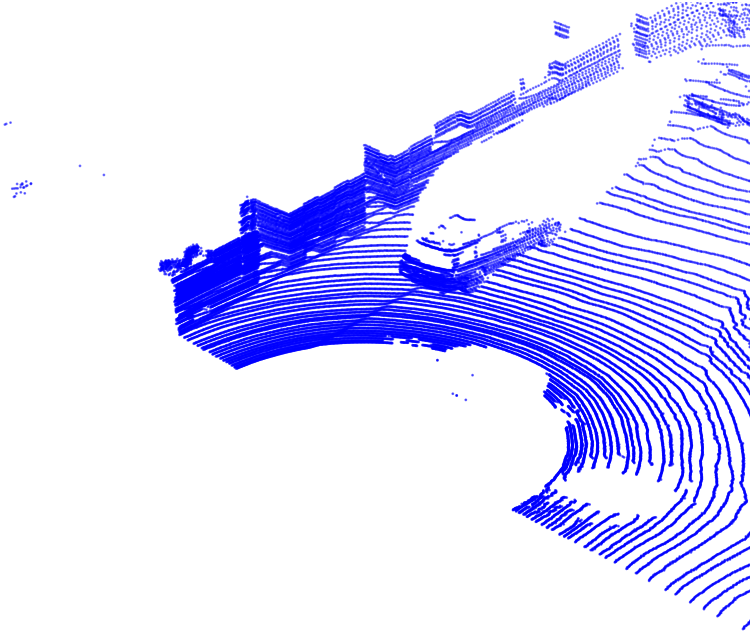}
		}		
		\subfloat[]{
			\includegraphics[width=0.45\columnwidth, clip]{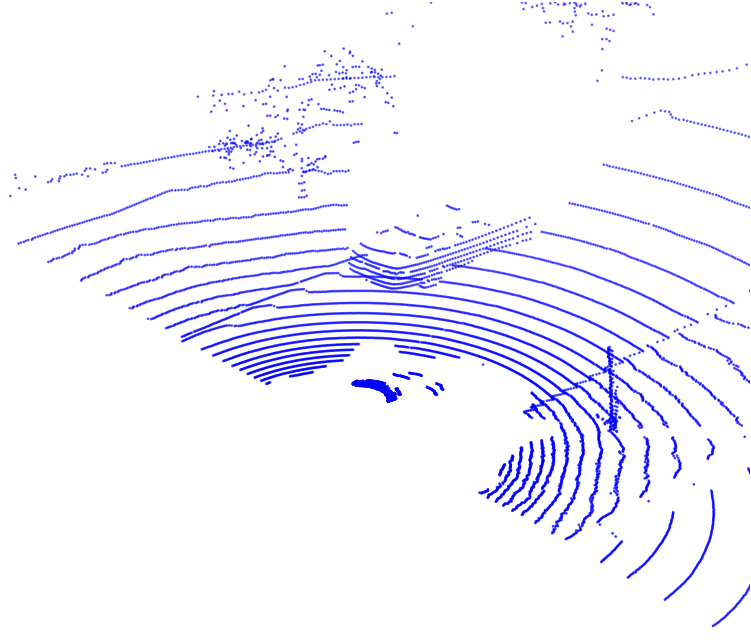} 
		}		
	\end{tabular}
	\caption{Example visualization showing the difference in sparsity level of LiDAR point clouds and 2D range view (RV) image projections. (a) and (c) visualize a point cloud and its generated 2D RV image projection from the SemanticKITTI dataset \cite{behley2019semantickitti}. (b) and (d) visualize a point cloud and its generated 2D RV image projection from the nuScenes dataset \cite{caesar2020nuscenes}. Best viewed in color with zoom.			
		\label{fig:range_img_differences} }
\end{figure}

Recently, deep learning approaches have shown to produce promising results for LiDAR semantic segmentation where the goal is to assign a class label to each point in the 3D LiDAR point cloud data. There exist LiDAR semantic segmentation approaches \cite{landrieu2018large,qi2017pointnet,qi2017pointnet++,zhu2021cylindrical} that directly operate on 3D LiDAR point clouds. There are also techniques that convert 3D point cloud to a different representation such as voxel \cite{zhou2018voxelnet} and image \cite{cortinhal2020salsanext} for further processing.  The methods \cite{cortinhal2020salsanext,milioto2019rangenet++,alonso20203d} where 3D LiDAR point clouds are projected onto a spherical surface to generate 2D range view (RV) images are especially appealing since they are suitable for training 2D convolution neural networks. However, much of the success of these deep learning methods is driven by the huge amount of labeled data that are required for supervision during the training. Moreover, it is not rare to see a well-performing existing model yield a significantly lower performance when it is trained on one dataset (\textit{source domain}, e.g., SemanticKITTI \cite{behley2019semantickitti}) and tested on another dataset (\textit{target domain}, e.g., nuScenes \cite{caesar2020nuscenes}). This scenario occurs due to the shift in the underlying distributions of the datasets. 3D point clouds suffer from this shift mainly due to variation in LiDAR sensors intrinsic/extrinsic parameters such as number of beams (sensor with more beams generate more dense point clouds data) and sensor placement (point clouds coordinates are relative to the sensor position) \cite{corralsoto_et_al_icra2021_lcp,Alonso2020DomainAI}. In the case of 2D RV images, this translates into dissimilarity in the density (or sparsity) of the 2D RV image samples on similar objects from the two different domains, which can be perceived as projection holes and missing lines as depicted in Fig. \ref{fig:range_img_differences} (a) and (b).

Domain adaptation techniques could potentially help address these issues and reduce the impact of domain shift. However, most of the advanced approaches \cite{sun2016deep,lee2019sliced,vu2019advent,wang2018deep} on domain adaption in semantic segmentation primarily focus on RGB images. In contrast to RGB images, there is lack research on domain adaptation in LiDAR semantic segmentation and research in this area is in its early stage. Therefore, in this work, we develop a domain adaptation strategy for LiDAR semantic segmentation. 

We propose an unsupervised domain adaptation (UDA) framework for LiDAR semantic segmentation. Specifically, we introduce three key modules to improve domain adaptation performance of a model on 3D LiDAR data from different domains. Firstly, we present a self-supervised auxiliary task that facilitates feature learning using unlabeled target domain LiDAR data. Secondly, we propose an unpaired mask transfer strategy to reduce the domain shift induced from difference in sparsity level of labeled source and unlabeled target LiDAR data. Lastly, we introduce light-weight gated adapter modules that are inserted in the network to capture target domain-specific knowledge. Note that, in this paper, we introduce these modules in a projection-based LiDAR semantic segmentation method (i.e., SalsaNext \cite{cortinhal2020salsanext}) since this method achieves state-of-the-art performance both in terms of speed and accuracy for LiDAR semantic segmentation, which is very crucial for autonomous driving systems. Nevertheless, we seek to improve its ability to perform well on data from different domains.

In summary, the contributions of this paper are as follows: (1) We propose a novel framework for unsupervised domain adaptation (UDA) in projection-based LiDAR semantic segmentation; (2) To bridge the domain gap between a labeled source domain and an unlabeled target domain, we propose three key modules: a self-supervised auxiliary task using target data, an unpaired mask transfer mechanism between source and target data, and the gated adapter module; and (3) We conduct extensive experiments adapting from both real-to-real and synthetic-to-real autonomous driving datasets to demonstrate the effectiveness of our approach.

\section{Related Work}

We aim to perform 3D semantic segmentation that assigns a semantic label to each point in 3D LiDAR point cloud data. Traditional methods use hand-crafted features from point cloud statistics information and geometric constraints \cite{xie2020linking}. Recent deep learning methods for 3D LiDAR semantic segmentation achieve promising results while operating mainly in two ways \cite{Alonso2020DomainAI}. First, some approaches \cite{landrieu2018large,qi2017pointnet,qi2017pointnet++,zhu2021cylindrical} directly work on 3D points by feeding raw point cloud as an input to the network. Second, other approaches \cite{alonso20203d,zhou2018voxelnet,cortinhal2020salsanext,wu2018squeezeseg} transform 3D point cloud data to a different representation (such as voxel and image) and then use it as an input. Recent methods  \cite{cortinhal2020salsanext,wu2018squeezeseg} that project the LiDAR point cloud onto a 2D image view are gaining popularity since they achieve superior performance and enable direct application of standard 2D convolutions. Our semantic segmentation backbone is based on a state-of-the-art projection-based LiDAR semantic segmentation network, SalsaNext \cite{cortinhal2020salsanext}, but we focus on improving its domain adaptation ability to 3D point cloud data from different LiDAR sensors.    

A major drawback of existing methods is that they often yield inferior performance when there is mismatch in the distribution of training and testing data. Unsupervised Domain Adaptation (UDA) models aim to handle this issue by revealing unlabeled test (target domain) data in addition to the labeled (source domain) data during the training. There is a dominant line of works (e.g., maximum mean discrepancy \cite{long2015learning}, adversarial training \cite{ganin2016domain,tzeng2017adversarial,hoffman2018cycada}, etc.) in UDA that follow the core idea of aligning the features from source and target domain such that they are domain-invariant. There is some work that minimizes the entropy \cite{vu2019advent} on output probabilities of the unlabeled target domain data. However, these prior methods focus on RGB-data, whereas we focus on UDA for LiDAR point cloud data where relatively little research has been done. Wu et al. \cite{wu2019squeezesegv2}, Qin et al. \cite{qin2019pointdan}, Jaritz et al. \cite{jaritz2020xmuda}, and Achituve et al. \cite{achituve2021self} perform domain adaptation for 3D point clouds but they do not explicitly focus on variations in 3D point cloud data due to difference in LiDAR sensor configurations.    

Our work is also related to the domain adaptation approaches that introduce the residual adapter module \cite{rebuffi2017_nips,rebuffi2018_cvpr} in an existing network to capture different domain information through a small number of domain-specific parameters. However, their application is currently limited to RGB images. In this paper, we study these residual adapters for domain adaptation in 3D point clouds data.

\section{Method}
\begin{figure*}[h] 
	\centering
	\includegraphics[width=1.0\textwidth, clip]{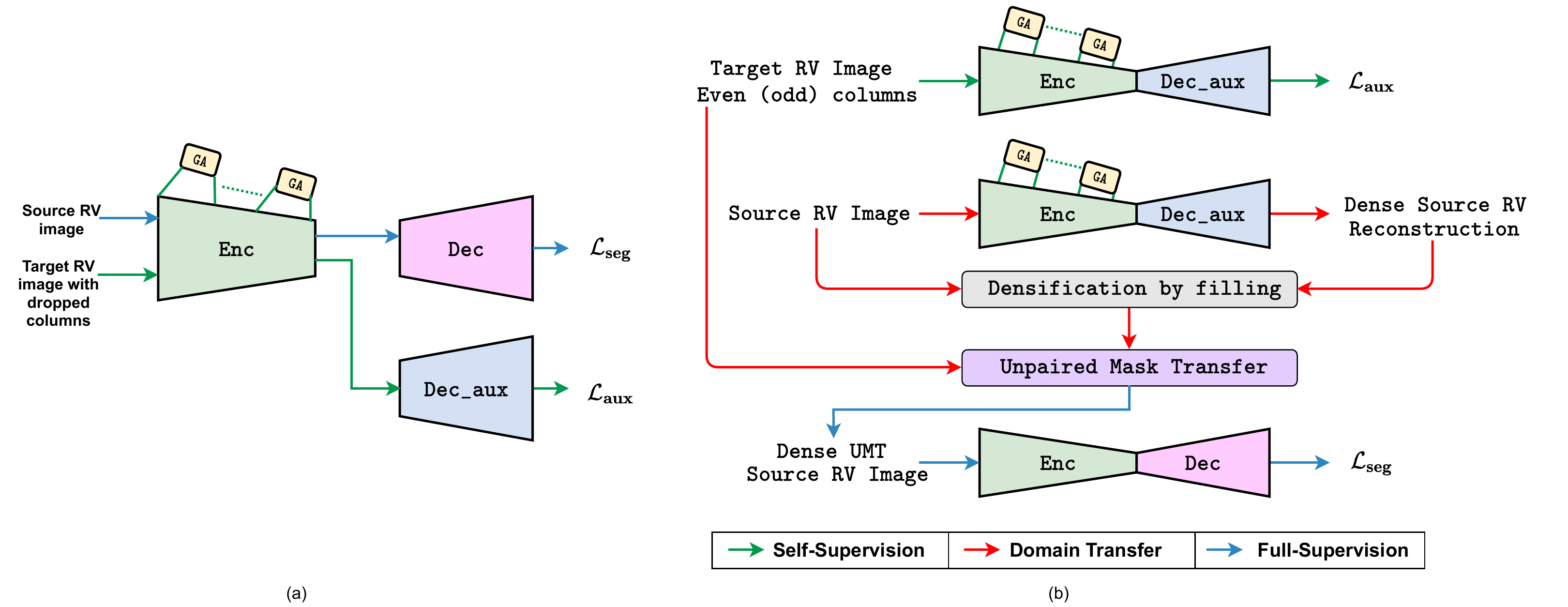} 
	\caption{Illustration of our UDA framework. (a) An overview of the proposed network architecture for UDA. The architecture consists of an encoder (\texttt{Enc}), gated adapter modules (\texttt{GA}), a primary decoder (\texttt{Dec}) that predicts logits for semantic segmentation, and an auxiliary decoder (\texttt{Dec\_aux}) responsible for self-supervision through the auxiliary task \textit{RVIC}. (b) An overview of our UDA training pipeline. Algorithm \ref{algo:train} presents the training steps (for one batch iteration) of our network.}
	\label{fig:overall}
\end{figure*}

Our goal is to learn an unsupervised domain adaptation (UDA) model for LiDAR semantic segmentation. In UDA, we have a source domain with a set of labeled LiDAR point clouds $\{x_i^s, y_i^s\}_{i=1}^S$ and a target domain with a set of unlabeled LiDAR point clouds $\{x_j^t\}_{j=1}^T$, where $x_i^s$ and $x_j^t$ denote the source and target 3D points, respectively, and $y_i^s \in Y = \{1,...,C\}$ indicate the semantic label within $C$ object classes for a source domain 3D point $x_i^s$. Note that the source and target domain point clouds are from different LiDAR sensors, and our aim to predict accurate semantic labels on test LiDAR points from the target domain.

We follow prior work \cite{cortinhal2020salsanext,milioto2019rangenet++} and project 3D LiDAR point cloud (from both source and target domain) onto a spherical surface that results in a 2D Range View (RV) image representation (example visualization in Fig. 1 (top)) suitable for standard convolution operations. We adopt this technique since prior research \cite{cortinhal2020salsanext} shows that projection-based approaches (such as 2D RV) achieve higher accuracy and run significantly faster than the methods that directly operate on raw 3D point clouds. Similar to the prior work \cite{cortinhal2020salsanext}, in the projection, we store 3D point coordinates, intensity and range values in separate RV image channels. In the end, we obtain $[h \times w \times 5]$ image representation for the point cloud, where $h$ and $w$ denote the height and width of the projected image, respectively. Note that this 2D RV image projection may contain many holes or empty pixels.

At the core of our UDA method is the introduction of three key modules: (i) a range view image completion task which is an auxiliary task for self-supervision from unlabeled target domain data, (ii) an unpaired mask transfer scheme between source and target domain data, and (iii) a gated adapter module to learn target domain-specific information. In the following, we firstly discuss each of these mechanisms in detail. Next, we present our network architecture and its training details.    

\subsection{Range View Image Completion}\label{sec:aux_task}
On the unlabeled target data (i.e., its RV image representation), we define a self-supervised auxiliary task, namely, \textit{Range View Image Completion (RVIC)}. From the input RV image, we drop the alternate columns and make the model predict these dropped columns as a regression problem. We find this is a simple and empirically effective task at feature learning using target data for the network.

This auxiliary task shares some of the model parameters $\bm{\theta}_{E}$ (i.e., \texttt{Enc}). It also has its own task-specific parameters $\bm{\theta}_{GA} = (\theta_{GA}^1,...,\theta_{GA}^n)$ and $\bm{\theta}_{A}$ (i.e., \texttt{Dec\_aux}). Fig. \ref{fig:overall}(a) shows the overall architecture and Sec. \ref{sec:arch} discusses its details. We define mean squared error as the loss function $\mathcal{L}_{aux}(x_j^t; \bm{\theta}_{E}, \bm{\theta}_{GA}, \bm{\theta}_{A})$ on this auxiliary task. 

\subsection{Unpaired Mask Transfer}

One of the main reasons for domain discrepancy in point clouds from different LiDAR sensors is the difference in their degree of sparsity. Different LiDAR sensors usually vary in their number of beams, and thereby do not capture the environment in a same way. A sensor with higher number of beams produce much denser point clouds as opposed to sensor with a lower number of beams. For example, Fig. \ref{fig:range_img_differences} (c) shows a point cloud captured from a 64-beam LiDAR sensor is denser than the point cloud captured from a 32-beam LiDAR sensor in Fig. \ref{fig:range_img_differences} (d). This difference in the level of sparsity is also reflected in their respective 2D RV images (Fig. \ref{fig:range_img_differences} (a) and (b)) in the form of projection holes. 

To tackle this issue, we propose an \textit{Unpaired Mask Transfer (UMT)} mechanism between source and target domain. The key idea is to match the level of sparsity of source RV images with the target RV images. Our UMT mechanism mainly consists of three steps. First, we use the RVIC mechanism on the source RV image to make it dense by filling its projection holes. Second, we randomly select a target RV image and generate its binary mask whose each entry indicate whether the corresponding pixel in the RV image has a value or not. Finally, we perform elementwise product between the target binary mask and the densified source RV image. This way we obtain a modified source RV image that is aligned with the target RV image in terms of level of its sparsity. Such an alignment between source and target would minimize the domain difference induced by variance in their degree of sparsity and hence would allow the network to perform well on different domains. Note that a source RV image can be randomly paired with target RV image in UMT, and thus the mechanism is unpaired. We also provide the steps involved in the UMT mechanism in Algorithm \ref{algo:train}.

\begin{algorithm*}[t]
	\SetAlgoLined
	\textbf{Arguments:} \vspace{-0.5ex}
	\begin{itemize}
		\item \texttt{model}: (\texttt{Enc}, \texttt{GA}, \texttt{Dec}, \texttt{Dec\_aux}) \vspace{-0.5ex}
		\item \texttt{image\_s}, \texttt{image\_t} : 2D RV image projection ($h \times w \times d$) of 3D point clouds for source (s) and target (t). \vspace{-0.5ex}
		\item \texttt{mask\_s}, \texttt{mask\_t} : 2D projection mask ($h \times w$) $\in[0, 1]$ indicating pixels containing valid projection. \vspace{-0.5ex}
		\item \texttt{label\_s} : 2D label projection ($h \times w$) corresponding to \texttt{image\_s} (source only).  \vspace{-0.5ex}
		\item $\mathcal{L}_{seg}$, $\mathcal{L}_{aux}$ : Fully supervised segmentation loss and self-supervised auxiliary loss, respectively. \vspace{-0.5ex}
	\end{itemize}
	
	\vspace{1ex}\textbf{\textcolor{PineGreen}{\# Step 1: Self-supervision (RVIC) step with target input}}\\
	\texttt{image\_t}, \texttt{image\_aux\_t} $\gets$ Split \texttt{image\_t} into even-odd vertical lines (columns) randomly \\
	\texttt{mask\_t}, \texttt{mask\_aux\_t} $\gets$ Split \texttt{mask\_t} similar to above \\
	\texttt{GA} $\gets$ \texttt{True} \textcolor{PineGreen}{\# Enable gated adapters}\\
	\texttt{model.train()}; \textcolor{PineGreen}{\# Set the model to training mode}\\
	\texttt{comp\_t} $\gets$ \texttt{model.Dec\_aux} ( \texttt{model.Enc\_and\_GA} ( image\_t ) ); \\
	\texttt{loss\_t} $\gets$ $\mathcal{L}_{aux}$( \texttt{comp\_t}, \texttt{image\_aux\_t}, \texttt{mask\_aux\_t} ); \\
	\texttt{loss\_t.backward()}; \textcolor{PineGreen}{\# Run the backward pass through \texttt{encoder} and \texttt{Dec\_aux}}\\ \vspace{1ex}
	
	\textbf{\textcolor{PineGreen}{\# Step 2 : Source RV image densification by filling its holes/missing pixels followed by UMT}}\\
	\texttt{model.eval()}; \textcolor{PineGreen}{\# Set the model to evaluation mode}\\
	\textcolor{PineGreen}{\# Obtain dense source RV image from the auxiliary decoder}\\
	\texttt{comp\_s} $\gets$ \texttt{model.Dec\_aux} ( \texttt{model.Enc\_and\_GA} ( image\_s ) ); \\
	\textcolor{PineGreen}{\# Fill in the holes/missing pixels in the source using the prediction from  above step}\\
	\texttt{mask\_inv\_s} $\gets$ 1 - \texttt{mask\_s}; \textcolor{PineGreen}{\# inverted mask with missing pixels == 1}\\
	\texttt{image\_s[mask\_inv\_s]} $\gets$ \texttt{comp\_s[mask\_inv\_s]}; \textcolor{PineGreen}{\# Filling the holes/missing pixels only}\\
	\textcolor{PineGreen}{\# Mask transfer from target to source}\\
	\texttt{mask\_t} $\gets$ \texttt{mask\_t} + \texttt{mask\_aux\_t};  \textcolor{PineGreen}{\# Combine the split masks again for target}\\
	\texttt{mask\_s} $\gets$ \texttt{mask\_s} $\odot$ \texttt{mask\_t};  \textcolor{PineGreen}{\# Transfer target mask to source with elementwise product}\\
	\texttt{label\_s} $\gets$ \texttt{label\_s} $\odot$ \texttt{mask\_s};  \textcolor{PineGreen}{\# Apply transferred mask on labels with elementwise product}\\
	\texttt{image\_s} $\gets$ \texttt{image\_s} $\odot$ \texttt{mask\_t};  \textcolor{PineGreen}{\# Transfer target mask on images with elementwise product and broadcasting}\\ \vspace{1ex}
	
	\textbf{\textcolor{PineGreen}{\# Step 3 : Full supervision with source input and its label}}\\
	\texttt{GA} $\gets$ \texttt{False} \textcolor{PineGreen}{   \# Disable gated adapters}\\
	\texttt{model.train()}; \textcolor{PineGreen}{\# Set the model to training mode}\\
	\textcolor{PineGreen}{\# Get the class prediction from mask transferred source RV image and compute supervised loss} \\
	\texttt{pred\_s} $\gets$ \texttt{model.Dec} ( \texttt{model.Enc} ( image\_s ) ); \\
	\textcolor{PineGreen}{\# Compute loss on the valid pixels from transferred source mask} \\
	\texttt{loss\_s} $\gets$ $\mathcal{L}_{seg}$( \texttt{pred\_s}, \texttt{labels\_s}, \texttt{mask\_s} ); \\ 
	\texttt{loss\_s.backward()}; \textcolor{PineGreen}{\# Run the backward pass through \texttt{encoder} and \texttt{decoder}}
	
	\caption{PyTorch style training procedure (one batch iteration) of our UDA network}
	\label{algo:train}
\end{algorithm*}

\subsection{Gated Adapter}
\begin{figure}[h]
	\begin{center}
		\includegraphics[width=0.48\textwidth]{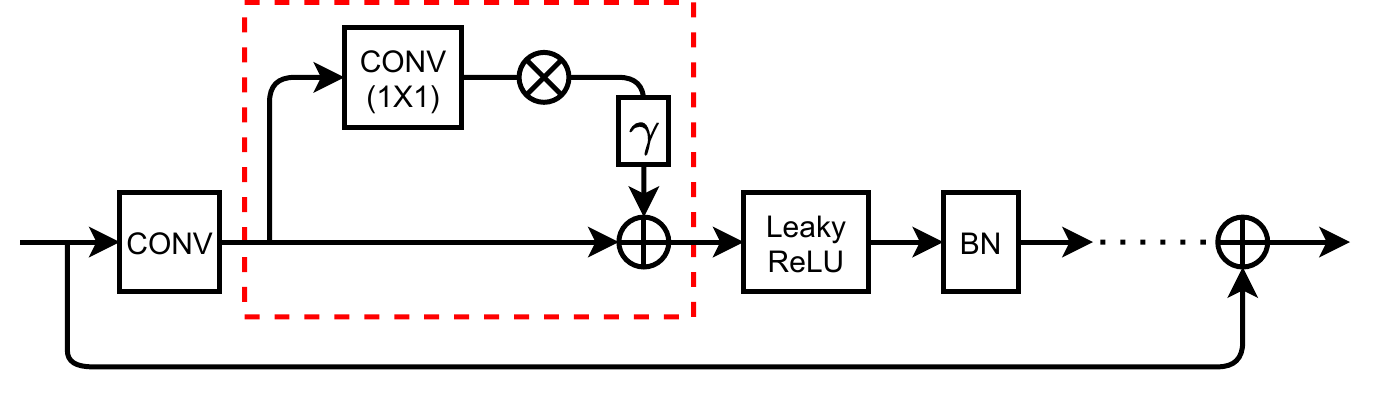}
	\end{center}
	\caption{The design of the gated adapter (\texttt{GA}) module (in red).}
	\label{fig:GA}
\end{figure}
In domain adaptation, there is a line of work that  adapts an existing network to a new domain with the help of a small number of domain-specific parameters that are attached to the network to account for domain differences \cite{rebuffi2017_nips,rebuffi2018_cvpr}. Rebuffi et al. \cite{rebuffi2017_nips} propose the residual adapter module that consists of additional parametric convolution layers to learn domain-specific information. We extend this idea and introduce the \textit{gated adapter (GA) module}. The incoming feature representation $x$ of an unlabeled target domain sample is firstly transformed by a light-weight $1 \times 1$ convolution operation $f(x;\alpha)$ with learnable parameters $\alpha$. Next, we introduce a gating mechanism $\gamma$, a learnable scalar which is initialized to 0 and is responsible for weighing the convolutional adjustments through $f(x;\alpha)$. We visualize a gated adapter module in Fig. \ref{fig:GA}. We express the operation within a gated adapter module as follows:
\begin{equation}\label{eqn:gated_adapter}
	g(x;\gamma,\alpha) = x + \gamma f(x;\alpha).
\end{equation}

The advantage of GA is that it can be plugged into any network and has a very small number learnable domain-specific parameters. The intuition behind learnable gate $\gamma$ (which we experimentally find performs better than when fixed to a value, e.g., $1$) is to allow the network to gradually learn to assign weight to target domain evidence, thereby regulate the contributions from the convolutional adjustments.

\subsection{Network Architecture}\label{sec:arch}

Our UDA network (see Fig. \ref{fig:overall})(a) is based on SalsaNext \cite{cortinhal2020salsanext}, which is a state-of-the-art encoder-decoder style LiDAR semantic segmentation network. We adapt SalsaNext for UDA and make a number of modifications to it. One change is that we insert gated adapter (\texttt{GA}) next to the convolution operations in each of the ResNet blocks \cite{he2016deep} of its encoder (\texttt{Enc}). Another change is that we introduce an auxiliary decoder (\texttt{Dec\_aux}) that is identical in architecture to the primary decoder (\texttt{Dec}) except the number of output channels in the last convolution layer. Specifically, we obtain the output of dimension $[h \times w \times C]$ from \texttt{Dec} where $C$ is the number of semantic classes, whereas the output of \texttt{Dec\_aux} is of the same dimension (i.e., $[h \times w \times 5]$) as input. For source, we feed its RV image as an input to the network. However, for target, we feed its RV image with dropped alternate columns to perform the RVIC task (see Sec. \ref{sec:aux_task}).

We denote the parameters in \texttt{Enc}, \texttt{GA}, \texttt{Dec}, and \texttt{Dec\_aux} by $\bm\theta_{E}, \bm\theta_{GA} = (\theta_{GA}^1,...,\theta_{GA}^n), \bm\theta_{D}$, and $\bm\theta_{A}$, respectively. We have $n$ ($=20$) \texttt{GA} modules in \texttt{Enc} (in each conv block of ResNet blocks) to capture target domain-specific information.

\subsection{Training Algorithm and Optimization}
Algorithm \ref{algo:train} describes the training procedure of our network. For simplicity, we present the algorithm for just one training batch iteration that includes a batch of labeled source data and a batch unlabeled target data. We also visualize our training pipeline in Fig. \ref{fig:overall}(b).

On the labeled source data, we compute the weighted cross-entropy loss (similar to \cite{cortinhal2020salsanext}) $\mathcal{L}_{seg}(x_i^s,y_i^s; \bm{\theta}_{E},\bm{\theta}_{D})$ on the prediction of \texttt{Dec}. On the unlabeled target data, we measure the auxiliary loss $\mathcal{L}_{aux}(x_j^t; \bm{\theta}_{E}, \bm{\theta}_{GA}, \bm{\theta}_{A})$ (see Sec. \ref{sec:aux_task}) on the prediction of \texttt{Dec\_aux}. 

The goal of our learning is to minimize the combination of these two losses and find the optimal parameters $\bm\theta_{E}^*, \bm\theta_{GA}^*, \bm\theta_{D}^*,$ and $\bm\theta_{A}^*$ in \texttt{Enc}, \texttt{GA}, \texttt{Dec}, and \texttt{Dec\_aux}, respectively. Therefore, we aim to solve:  
\begin{eqnarray}\label{eqn:loss}\nonumber
	\bm\theta_{E}^*, \bm\theta_{GA}^*, \bm\theta_{D}^*, \bm\theta_{A}^*&=&  \argmin_{\bm{\theta}_{E},\bm{\theta}_{GA},\bm{\theta}_{D},\bm{\theta}_{A}} \big(\mathcal{L}_{seg}(x_i^s,y_i^s; \bm{\theta}_{E},\bm{\theta}_{D}) \\ 
	&&  + \lambda\mathcal{L}_{aux}(x_j^t; \bm{\theta}_{E}, \bm{\theta}_{GA}, \bm{\theta}_{A}) \big),
\end{eqnarray}
where $\lambda$ is a hyperparameter set to $1e$-$6$  to control the relative importance of the auxiliary loss $\mathcal{L}_{aux}$.

\subsection{Evaluation and Post-processing}
In inference mode, we do not require \texttt{Dec\_aux}. In other words, we forward the RV image of a 3D point cloud from the test set of target domain dataset to the trained \texttt{Enc}, \texttt{GA} and \texttt{Dec} to obtain the network prediction.  

A drawback of RV image-based projection representation is that multiple LiDAR points may get projected to same image pixel. This causes problem when the RV image is projected back to the original 3D point cloud space. To cope with this issue, we follow prior work \cite{cortinhal2020salsanext,milioto2019rangenet++} and perform the kNN-based post-processing on the network output. Note that we employ this post-processing only during evaluation on the target domain samples and not during training.

\section{Experiments}

\subsection{Settings}\label{sec:settings}

{\bf Datasets.} We experiment adapting from real-to-real and synthetic-to-real scenarios for LiDAR semantic segmentation. The evaluation on synthetic-to-real scenario is very appealing since it is much easier to collect synthetic labeled dataset as opposed to the real labeled dataset. We experiment with nuScenes \cite{caesar2020nuscenes} and SemanticKITTI \cite{behley2019semantickitti} for real-to-real adaptation and GTA-LiDAR \cite{wu2019squeezesegv2} and real KITTI \cite{geiger2012we} for synthetic-to-real adaptation.

nuScenes \cite{caesar2020nuscenes} is a large-scale LiDAR point cloud segmentation dataset with 28,130 and 6,019 samples in training and validation sets, respectively. This dataset is captured using a 32-beam LiDAR sensor as opposed to 64-beam LiDAR for SemanticKITTI. Officially, this dataset comes with annotation for 16 semantic categories. However, to pair with SemanticKITTI for domain adaptation, we merge \{Bus, Construction-vehicle, Ambulance, Police-vehicle, Trailer\} into the Other-vehicle category.

SemanticKITTI \cite{behley2019semantickitti} is a dataset created using 64-beam Velodyne HDL-64E laser scanner. Following \cite{behley2019semantickitti}, we use sequence-\{0-7, 9-10\} (19,130 scans) for training and sequence-08 (4,071 scans) for evaluation. To pair with nuScenes for domain adaptation, we firstly use the 11 intersecting semantic classes between them. Additionally, we also rename and merge some of the semantic classes to pair with nuScenes for domain adaptation. In particular, \{Bicycle, Bicyclist\} and \{Motorcycle, Motorcyclist\} are merged into the Bicycle and Motorcycle categories, respectively. Person and Road are renamed to Pedestrian and Drivable\_surface, respectively. \{Vegetation, Trunk\} are merged into Vegetation. \{Building, Fence, Other-structure, Pole, Traffic-sign\} are merged into Manmade. Note that SemanticKITTI has much more densely annotated scans (nearly $100K$ points per scan) than nuScenes (nearly $30K$ points per scan).  

GTA-LiDAR \cite{wu2019squeezesegv2} is a synthetic dataset with only two categories (i.e., car and pedestrian). There are 121,087 scans in range image projection format. In our experiments, we use the last 9,087 samples for validation and the remaining for training.

KITTI \cite{wu2018squeezeseg,geiger2012we} has 8,057 and 2,791 samples in training and testing, respectively. It has three semantic categories: Car, Pedestrian, and Cyclist. However, to pair with GTA-LiDAR, we ignore Cyclist category that results in 7,930 samples for training and 2,653 samples for testing \cite{zhao2020epointda}.

{\bf Evaluation Metric.} Following prior work  \cite{cortinhal2020salsanext} in LiDAR semantic segmentation, we use mean intersection over union (mIoU) as the evaluation metric that is given by $mIoU = \frac{1}{C}\sum_{c=1}^{C}\frac{|P_c \cap G_c|}{|P_c \cup G_c|}$, where $P_c$ denote the set of points with the class prediction $c$, $G_c$ denote the ground-truth point set for class $c$ and $|\cdot|$ denote the set's cardinality.

{\bf Implementation details.} Our method is implemented using PyTorch with the same hyper-parameters as of SalsaNext \cite{cortinhal2020salsanext}. We employ the stochastic gradient descent (SGD) with a warmup scheduler as our optimizer. The initial learning rate, momentum, and weight decay are set to $0.01$, $0.9$, and $0.0001$, respectively. We fix the batch size to $24$ for semanticKITTI and nuScenes, and $48$ for GTA-LiDAR and KITTI. We train our models on four \texttt{NVIDIA Tesla V100 32GB} GPUs. Note that following the SemanticKITTI \cite{behley2019semantickitti} benchmark, we ignore the background class while training our real-to-real experiments, which is not the case for the synthetic-to-real setup \cite{wu2019squeezesegv2}. We use the weighted Cross-Entropy error as our supervised loss function, where the weight per class is set to the square root of the reciprocal class-frequency \cite{cortinhal2020salsanext}.

\subsection{Evaluation Results for UDA}

\begin{table*}[ht]
	\caption{Comparison (IoU\% for each class and mIoU\%) with the state-of-the-art UDA methods for LiDAR semantic segmentation from SemanticKITTI (semKITTI) to nuScenes and nuScenes to semKITTI. We also include the performance of fully-supervised method (Supervised) to indicate the upper-bound on the datasets.}
	\begin{threeparttable}
		\centering
		\renewcommand{\arraystretch}{1.3}
		\adjustbox{max width=\textwidth}{%
			\begin{tabular}{c|c|c|ccccccccccc|c}
				\hline
				\rotatebox{0}{Source} & 
				\rotatebox{0}{Target} & 
				\rotatebox{0}{Method} &
				\rotatebox{90}{Car} &
				\rotatebox{90}{Bicycle} &
				\rotatebox{90}{Motorcycle} &
				\rotatebox{90}{Other\_vehicle} &
				\rotatebox{90}{Pedestrian} &
				\rotatebox{90}{Truck} &
				\rotatebox{90}{Drivable\_surface} &
				\rotatebox{90}{Sidewalk} &
				\rotatebox{90}{Terrain} &
				\rotatebox{90}{Vegetation} &
				\rotatebox{90}{Manmade} &
				\multicolumn{1}{|c}{\rotatebox{0}{mIoU}} \\ \hline
				nuScenes & nuScenes & \textcolor{black}{Supervised} & 84.5 & 22.7 & 66.4 & 63.3 & 59.5 & 72.7 & 96.4 & 73.4 & 74.0 & 85.4 & 87.9 & 71.5 \\ \hdashline
				\multirow{7}{*}{semKITTI} & \multirow{7}{*}{nuScenes} & Na\"ive  & 35.7 & 0.2 & 0.4 & 5.7 & 7.5 & 8.1 & 73.8 & 15.0 & 14.9 & 8.3 & 51.4 & 20.1 \\
				&  & CORAL\cite{sun2016deep}  & 51.0 & 0.9 & 6.0 & 4.0 & 25.9 & 29.9 & 82.6 & 27.1 & 27.0 & 55.3 & 56.7 & 33.3 \\
				&  & MEnt \cite{vu2019advent}  & 57.4 & 2.2 & 4.6 & 6.4 & 22.6 & 19.3 & 82.3 & 28.8 & 29.9 & 46.8 & 64.2 & 33.1 \\
				&  & AEnt \cite{vu2019advent} & 57.4 & 1.1 & 8.6 & 6.7 & 24.0 & 10.1 & 81.0 & 25.4 & 26.6 & 34.2 & 58.9  & 30.4 \\
				&  & (M+A)Ent \cite{vu2019advent} & 57.3 & 1.1 & 2.3 & 6.8 & 23.4 & 7.9 & 83.5 & 32.6 & 31.8 & 43.3 & 62.3 & 32.0 \\
				&  & SWD \cite{lee2019sliced} & 45.3 & 2.1 & 2.2 & 3.4 & 25.9 & 10.6 & 80.7 & 26.5 & 30.1 & 43.9 & 60.2 & 30.1 \\
				&  & Ours & 54.4 & 3.0 & 1.9 & 7.6 & 27.7 & 15.8 & 82.2 & 29.6 & 34.0 & 57.9 & 65.7 & \textbf{34.5} \\			
				\hline 
				\hline
				semKITTI & semKITTI & \textcolor{black}{Supervised}  & 92.2 & 52.6 & 47.8 & 48.3 & 53.7 & 80.2 & 94.6 & 82.5 & 70.6 & 85.9 & 86.8 & 72.3 \\ \hdashline
				\multirow{7}{*}{nuScenes} & \multirow{7}{*}{semKITTI} & Na\"ive  & 7.7 & 0.1 & 0.9 & 0.6 & 6.4 & 0.4 & 30.4 & 5.7 & 28.4 & 27.8 & 30.2 & 12.6 \\
				&  & CORAL \cite{sun2016deep}  & 47.3 & 10.4 & 6.9 & 5.1 & 10.8 & 0.7 & 24.8 & 13.8 & 31.7 & 58.8 & 45.5 & 23.2 \\ 
				&  & MEnt \cite{vu2019advent}  & 27.1 & 2.0 & 2.3 & 3.4 & 9.5 & 0.4 & 29.3 & 11.3 & 28.0 & 35.8 & 39.0 & 17.1 \\
				&  & AEnt \cite{vu2019advent}  & 42.4 & 4.5 & 6.9 & 2.8 & 6.7 & 0.7 & 16.1 & 7.0 & 26.1 & 46.1 & 42.0 & 18.3 \\ 
				&  & (M+A)Ent \cite{vu2019advent} & 49.6 & 5.9 & 4.3 & 6.4 & 9.6 & 2.6 & 22.5 & 12.7 & 30.3 & 57.4 & 49.1 & 22.8 \\  
				&  & SWD \cite{lee2019sliced}  & 34.2 & 2.7 & 1.5 & 2.0 & 5.3 & 0.9 & 28.8 & 20.5 & 28.3 & 38.2 & 36.7 & 18.1 \\
				&  & Ours  & 49.6 & 4.6 & 6.3 & 2.0 & 12.5 & 1.8 & 25.2 & 25.2 & 42.3 & 43.4 & 45.3 & \textbf{23.5} \\
				
				\hline
			\end{tabular}%
		}
	\end{threeparttable}
	\label{table:kitti-nuscenes}
\end{table*}

\begin{table}[h!]
	\centering
	\caption{Comparison (IoU\% for each class and mIoU\%) for UDA from synthetic GTA-LiDAR dataset to real KITTI.}
	\begin{threeparttable}
		\renewcommand{\arraystretch}{1.3}
		\adjustbox{max width=0.6\textwidth}{%
			\begin{tabular}{c|c|c|cc|c}
				\hline
				\rotatebox{0}{Source} & 
				\rotatebox{0}{Target} & 
				\rotatebox{0}{Method} &
				\rotatebox{0}{Car} &
				\rotatebox{0}{Pedestrian} &
				\multicolumn{1}{c}{\rotatebox{0}{mIoU}} \\ \hline
				KITTI & KITTI & \textcolor{black}{Supervised} & 66.9 & 28.0 & \multicolumn{1}{c}{47.4} \\ \hdashline
				\multirow{7}{*}{GTA-LiDAR} & \multirow{7}{*}{KITTI} & Na\"ive  & 17.5 & 14.9 & 16.2  \\
				&  & CORAL \cite{sun2016deep}  & 34.8 & 22.4 & 28.6 \\   
				&  & MEnt \cite{vu2019advent}  & 37.9 & 21.8 & 30.0 \\
				&  & AEnt \cite{vu2019advent}  & 34.3 & 23.0 & 28.6 \\ 
				&  & (M+A)Ent \cite{vu2019advent} & 29.8 & 22.5 & 26.1 \\  
				&  & SWD \cite{lee2019sliced} & 32.1 & 29.6 & 30.8 \\
				&  & Ours & 51.0 & 29.3 & \textbf{40.2} \\
				\hline
			\end{tabular}%
		}
	\end{threeparttable}
	\label{table:gta-kitti}
\end{table}

We examine the domain adaptation ability of our method on SemanticKITTI, nuScenes, GTA-LiDAR, and KITTI. Since there is lack of research on projection-based domain adaptation in 3D LiDAR point clouds, we compare with state-of-the-art UDA methods in 2D semantic segmentation. Specifically, we compare with CORAL \cite{sun2016deep}, Advent \cite{vu2019advent}, and SWD \cite{lee2019sliced}. We also compare with a baseline (we refer as Na\"ive) that directly evaluates a pre-trained model from source domain on target domain. For fair comparison, we re-implement the baseline methods with SalsaNext \cite{cortinhal2020salsanext} as their backbone network. Tables \ref{table:kitti-nuscenes} and \ref{table:gta-kitti} show the results. 

From Tables \ref{table:kitti-nuscenes} and \ref{table:gta-kitti}, we see a huge performance drop when we directly test the network trained on one domain to another (i.e., Na\"ive method). This highlights the importance of domain adaptation in LiDAR semantic segmentation. However, our method improves the performance while outperforming the prior arts. We observe performance gain by our method on  both the real-to-real (SemanticKITTI to nuScenes and vice-versa) and synthetic-to-real (GTA-LiDAR to KITTI) datasets.

\begin{figure}[h] 
	\centering
	\includegraphics[width=0.48\textwidth]{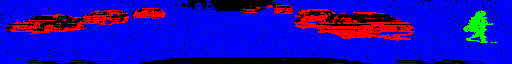}\vspace{0.7ex}
	\includegraphics[width=0.48\textwidth]{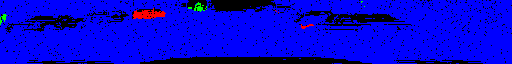}\vspace{0.7ex}
	\includegraphics[width=0.48\textwidth]{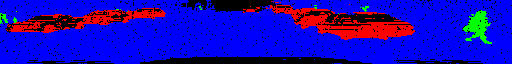}
	\caption{Example semantic segmentation results on  KITTI. Top, middle, and bottom rows show the ground-truth, the prediction of baseline with no adaptation (Na\"ive), and the prediction from our UDA framework, respectively. For better view, we show results on 2D RV image projections where each color denote a different semantic class.}
	\label{fig:qual}
\end{figure}

Fig. \ref{fig:qual} shows a few qualitative results when adapting from synthetic GTA-LiDAR to KITTI. We can see the improvement in the prediction from our UDA method in comparison to the baseline method.

\subsection{Ablation Study}
\begin{table}[h]
	\centering
	\caption{Ablation study on different modules in our method. sK, N, GTA denote the SemanticKITTI, nuScenes and GTA-LiDAR dataset, respectively. We report mIoU (\%) in each cell. The last row indicates our final UDA method.}
		\begin{tabular}{ c|c|c|c } 
			\hline
			Method & sK $\rightarrow$ N & N $\rightarrow$ sK & GTA $\rightarrow$ K \\
			\hline
			RVIC & 23.5 & 17.5 & 17.1  \\ 
			RVIC+UMT & 32.0 & 22.5 & 39.3  \\ 
			RVIC+UMT+GA & \textbf{34.5} & \textbf{23.5} & \textbf{40.2}  \\  
			\hline
		\end{tabular}
	\label{table:ablation}
\end{table}
We conduct an ablation study to incrementally examine the contribution of various modules in our UDA method. Table \ref{table:ablation} shows that as we introduce modules in our method there is improvement in the mIoU measure. This study demonstrates that each module contributes to the UDA task.

\section{Conclusion}
In this paper, we proposed an unsupervised domain adaptation (UDA) framework for LiDAR semantic segmentation. Our pipeline consists of a self-supervised auxiliary task that is coupled with an unpaired mask transfer mechanism to alleviate the impact of domain gap when adapting the model from source to target domain. Additionally, we also introduced light-weight gated adapter modules to inject target domain-specific evidence into the network. Extensive experiments on adapting from both real-to-real and synthetic-to-real datasets demonstrated our approach achieves state-of-the-art results.

\bibliographystyle{IEEEtran}
\bibliography{IEEEabrv,udass}

\begin{thebibliography}{10}
\providecommand{\url}[1]{#1}
\csname url@rmstyle\endcsname
\providecommand{\newblock}{\relax}
\providecommand{\bibinfo}[2]{#2}
\providecommand\BIBentrySTDinterwordspacing{\spaceskip=0pt\relax}
\providecommand\BIBentryALTinterwordstretchfactor{4}
\providecommand\BIBentryALTinterwordspacing{\spaceskip=\fontdimen2\font plus
\BIBentryALTinterwordstretchfactor\fontdimen3\font minus
  \fontdimen4\font\relax}
\providecommand\BIBforeignlanguage[2]{{%
\expandafter\ifx\csname l@#1\endcsname\relax
\typeout{** WARNING: IEEEtran.bst: No hyphenation pattern has been}%
\typeout{** loaded for the language `#1'. Using the pattern for}%
\typeout{** the default language instead.}%
\else
\language=\csname l@#1\endcsname
\fi
#2}}

\bibitem{landrieu2018large}
L.~Landrieu and M.~Simonovsky, ``Large-scale point cloud semantic segmentation
  with superpoint graphs,'' in \emph{IEEE Conference on Computer Vision and
  Pattern Recognition}, 2018.

\bibitem{qi2017pointnet}
C.~R. Qi, H.~Su, K.~Mo, and L.~J. Guibas, ``Pointnet: Deep learning on point
  sets for 3d classification and segmentation,'' in \emph{IEEE Conference on
  Computer Vision and Pattern Recognition}, 2017.

\bibitem{qi2017pointnet++}
C.~R. Qi, L.~Yi, H.~Su, and L.~J. Guibas, ``Pointnet++: Deep hierarchical
  feature learning on point sets in a metric space,'' \emph{arXiv preprint
  arXiv:1706.02413}, 2017.

\bibitem{zhu2021cylindrical}
X.~Zhu, H.~Zhou, T.~Wang, F.~Hong, Y.~Ma, W.~Li, H.~Li, and D.~Lin,
  ``Cylindrical and asymmetrical 3d convolution networks for lidar
  segmentation,'' in \emph{IEEE/CVF Conference on Computer Vision and Pattern
  Recognition}, 2021.

\bibitem{zhou2018voxelnet}
Y.~Zhou and O.~Tuzel, ``Voxelnet: End-to-end learning for point cloud based 3d
  object detection,'' in \emph{IEEE Conference on Computer Vision and Pattern
  Recognition}, 2018.

\bibitem{cortinhal2020salsanext}
T.~Cortinhal, G.~Tzelepis, and E.~E. Aksoy, ``Salsanext: Fast,
  uncertainty-aware semantic segmentation of lidar point clouds,'' in
  \emph{International Symposium on Visual Computing}, 2020.

\bibitem{milioto2019rangenet++}
A.~Milioto, I.~Vizzo, J.~Behley, and C.~Stachniss, ``Rangenet++: Fast and
  accurate lidar semantic segmentation,'' in \emph{IEEE/RSJ International
  Conference on Intelligent Robots and Systems}, 2019.

\bibitem{alonso20203d}
I.~Alonso, L.~Riazuelo, L.~Montesano, and A.~C. Murillo, ``3d-mininet: Learning
  a 2d representation from point clouds for fast and efficient 3d lidar
  semantic segmentation,'' \emph{IEEE Robotics and Automation Letters}, vol.~5,
  no.~4, pp. 5432--5439, 2020.

\bibitem{behley2019semantickitti}
J.~Behley, M.~Garbade, A.~Milioto, J.~Quenzel, S.~Behnke, C.~Stachniss, and
  J.~Gall, ``Semantickitti: A dataset for semantic scene understanding of lidar
  sequences,'' in \emph{IEEE/CVF International Conference on Computer Vision},
  2019.

\bibitem{caesar2020nuscenes}
H.~Caesar, V.~Bankiti, A.~H. Lang, S.~Vora, V.~E. Liong, Q.~Xu, A.~Krishnan,
  Y.~Pan, G.~Baldan, and O.~Beijbom, ``nuscenes: A multimodal dataset for
  autonomous driving,'' in \emph{IEEE/CVF Conference on Computer Vision and
  Pattern Recognition}, 2020.

\bibitem{corralsoto_et_al_icra2021_lcp}
E.~R. Corral-Soto, A.~Nabatchian, M.~Gerdzhev, and L.~Bingbing, ``Lidar
  few-shot domain adaptation via integrated cyclegan and 3d object detector
  with joint learning delay,'' in \emph{International Conference on Robotics
  and Automation}, 2021.

\bibitem{Alonso2020DomainAI}
I.~Alonso, L.~Riazuelo, L.~Montesano, and A.~C. Murillo, ``Domain adaptation in
  lidar semantic segmentation,'' \emph{arXiv preprint arXiv:2010.12239}, 2020.

\bibitem{sun2016deep}
B.~Sun and K.~Saenko, ``Deep coral: Correlation alignment for deep domain
  adaptation,'' in \emph{European Conference on Computer Vision}, 2016.

\bibitem{lee2019sliced}
C.-Y. Lee, T.~Batra, M.~H. Baig, and D.~Ulbricht, ``Sliced wasserstein
  discrepancy for unsupervised domain adaptation,'' in \emph{IEEE/CVF
  Conference on Computer Vision and Pattern Recognition}, 2019.

\bibitem{vu2019advent}
T.-H. Vu, H.~Jain, M.~Bucher, M.~Cord, and P.~P{\'e}rez, ``Advent: Adversarial
  entropy minimization for domain adaptation in semantic segmentation,'' in
  \emph{IEEE/CVF Conference on Computer Vision and Pattern Recognition}, 2019.

\bibitem{wang2018deep}
M.~Wang and W.~Deng, ``Deep visual domain adaptation: A survey,''
  \emph{Neurocomputing}, vol. 312, pp. 135--153, 2018.

\bibitem{xie2020linking}
Y.~Xie, J.~Tian, and X.~X. Zhu, ``Linking points with labels in 3d: A review of
  point cloud semantic segmentation,'' \emph{IEEE Geoscience and Remote Sensing
  Magazine}, vol.~8, no.~4, pp. 38--59, 2020.

\bibitem{wu2018squeezeseg}
B.~Wu, A.~Wan, X.~Yue, and K.~Keutzer, ``Squeezeseg: Convolutional neural nets
  with recurrent crf for real-time road-object segmentation from 3d lidar point
  cloud,'' in \emph{IEEE International Conference on Robotics and Automation},
  2018.

\bibitem{long2015learning}
M.~Long, Y.~Cao, J.~Wang, and M.~Jordan, ``Learning transferable features with
  deep adaptation networks,'' in \emph{International Conference on Machine
  Learning}, 2015.

\bibitem{ganin2016domain}
Y.~Ganin, E.~Ustinova, H.~Ajakan, P.~Germain, H.~Larochelle, F.~Laviolette,
  M.~Marchand, and V.~Lempitsky, ``Domain-adversarial training of neural
  networks,'' \emph{Journal of Machine Learning Research}, vol.~17, no.~1, pp.
  2096--2030, 2016.

\bibitem{tzeng2017adversarial}
E.~Tzeng, J.~Hoffman, K.~Saenko, and T.~Darrell, ``Adversarial discriminative
  domain adaptation,'' in \emph{IEEE Conference on Computer Vision and Pattern
  Recognition}, 2017.

\bibitem{hoffman2018cycada}
J.~Hoffman, E.~Tzeng, T.~Park, J.-Y. Zhu, P.~Isola, K.~Saenko, A.~Efros, and
  T.~Darrell, ``Cycada: Cycle-consistent adversarial domain adaptation,'' in
  \emph{International Conference on Machine Learning}, 2018.

\bibitem{wu2019squeezesegv2}
B.~Wu, X.~Zhou, S.~Zhao, X.~Yue, and K.~Keutzer, ``Squeezesegv2: Improved model
  structure and unsupervised domain adaptation for road-object segmentation
  from a lidar point cloud,'' in \emph{International Conference on Robotics and
  Automation}, 2019.

\bibitem{qin2019pointdan}
C.~Qin, H.~You, L.~Wang, C.-C.~J. Kuo, and Y.~Fu, ``Pointdan: A multi-scale 3d
  domain adaption network for point cloud representation,'' in \emph{Advances
  in Neural Information Processing Systems}, 2019.

\bibitem{jaritz2020xmuda}
M.~Jaritz, T.-H. Vu, R.~d. Charette, E.~Wirbel, and P.~P{\'e}rez, ``xmuda:
  Cross-modal unsupervised domain adaptation for 3d semantic segmentation,'' in
  \emph{IEEE/CVF Conference on Computer Vision and Pattern Recognition}, 2020.

\bibitem{achituve2021self}
I.~Achituve, H.~Maron, and G.~Chechik, ``Self-supervised learning for domain
  adaptation on point clouds,'' in \emph{IEEE/CVF Winter Conference on
  Applications of Computer Vision}, 2021.

\bibitem{rebuffi2017_nips}
S.-A. Rebuffi, H.~Bilen, and A.~Vedaldi, ``Learning multiple visual domains
  with residual adapters,'' in \emph{Advances in Neural Information Processing
  Systems}, 2017.

\bibitem{rebuffi2018_cvpr}
------, ``Efficient parametrization of multi-domain deep neural networks,'' in
  \emph{IEEE Conference on Computer Vision and Pattern Recognition}, 2018.

\bibitem{he2016deep}
K.~He, X.~Zhang, S.~Ren, and J.~Sun, ``Deep residual learning for image
  recognition,'' in \emph{IEEE Conference on Computer Vision and Pattern
  Recognition}, 2016.

\bibitem{geiger2012we}
A.~Geiger, P.~Lenz, and R.~Urtasun, ``Are we ready for autonomous driving? the
  kitti vision benchmark suite,'' in \emph{IEEE Conference on Computer Vision
  and Pattern Recognition}, 2012.

\bibitem{zhao2020epointda}
S.~Zhao, Y.~Wang, B.~Li, B.~Wu, Y.~Gao, P.~Xu, T.~Darrell, and K.~Keutzer,
  ``epointda: An end-to-end simulation-to-real domain adaptation framework for
  lidar point cloud segmentation,'' \emph{arXiv preprint arXiv:2009.03456},
  2020.

\end{thebibliography}

\end{document}